\documentclass[runningheads]{llncs}
\usepackage{graphicx}
\usepackage{
cite}
\usepackage{amsmath,amssymb,amsfonts}
\usepackage[colorlinks=true, citecolor=blue, linkcolor=blue]{hyperref}
\usepackage{amsmath,amssymb,amsfonts}
\usepackage{algorithmic}
\usepackage{graphicx}
\usepackage{textcomp}
\usepackage{multirow}
\usepackage{amsmath}
\usepackage{rotating}
\usepackage[misc]{ifsym}
\usepackage{amssymb}
\usepackage{xcolor}
 \usepackage{booktabs}

\newcommand{\bC}{\mathbf{C}}
\newcommand{\bD}{\mathbf{D}}

\newcommand{\bF}{\mathbf{F}} %

\newcommand{\bI}{\mathbf{I}}

\newcommand{\bK}{\mathbf{K}}

\newcommand{\bO}{\mathbf{O}}

\newcommand{\bR}{\mathbf{R}}
\newcommand{\bS}{\mathbf{S}}
\newcommand{\bT}{\mathbf{T}}
\newcommand{\bu}{\mathbf{u}}

\newcommand{\bX}{\mathbf{X}}

\newcommand{\nR}{\mathbb{R}}

\makeatletter
\DeclareRobustCommand\onedot{\futurelet\@let@token\@onedot}
\def\@onedot{\ifx\@let@token.\else.\null\fi\xspace}

\makeatother

\begin{document}
\title{Endo3R: Unified Online Reconstruction from Dynamic Monocular Endoscopic Video}

\titlerunning{Endo3R: Unified Online Reconstruction}

\author{Jiaxin Guo\inst{1} \and
Wenzhen Dong\inst{2} \and Tianyu Huang\inst{1,2} \and Hao Ding\inst{3} \and Ziyi Wang\inst{1} \and Haomin Kuang \inst{4} \and Qi Dou \inst{1} \and Yun-hui Liu\inst{1,2}\textsuperscript{(\Letter)}}  
\authorrunning{J. Guo et al.}
\institute{The Chinese University of Hong Kong, Hong Kong SAR, China  \and
Hong Kong Center for Logistics Robotics, Hong Kong SAR, China \and
Johns Hopkins University, Baltimore MD 21218, USA \and
Shanghai Jiao Tong University, Shanghai, China
}

\maketitle              %
\begin{abstract}
Reconstructing 3D scenes from monocular surgical videos can enhance surgeon's perception and therefore plays a vital role in various computer-assisted surgery tasks. However, achieving scale-consistent reconstruction remains an open challenge due to inherent issues in endoscopic videos, such as dynamic deformations and textureless surfaces. Despite recent advances, current methods either rely on calibration or instrument priors to estimate scale, or employ SfM-like multi-stage pipelines, leading to error accumulation and requiring offline optimization. In this paper, we present Endo3R, a unified 3D foundation model for online scale-consistent reconstruction from monocular surgical video, without any priors or extra optimization. Our model unifies the tasks by predicting globally aligned pointmaps, scale-consistent video depths, and camera parameters without any offline optimization. The core contribution of our method is expanding the capability of the recent pairwise reconstruction model to long-term incremental dynamic reconstruction by an uncertainty-aware dual memory mechanism. 
The mechanism maintains history tokens of both short-term dynamics and long-term spatial consistency. Notably, to tackle the highly dynamic nature of surgical scenes, we measure the uncertainty of tokens via Sampson distance and filter out tokens with high uncertainty. Regarding the scarcity of endoscopic datasets with ground-truth depth and camera poses, we further devise a self-supervised mechanism with a novel dynamics-aware flow loss. Abundant experiments on SCARED and Hamlyn datasets demonstrate our superior performance in zero-shot surgical video depth prediction and camera pose estimation with online efficiency. Project page: \url{https://wrld.github.io/Endo3R/}.

\end{abstract}
\section{Introduction}

Reconstructing surgical scenes from endoscopic videos is crucial for minimally invasive surgery, benefiting various downstream tasks including surgical planning, intraoperative navigation, and robotic surgical automation~\cite{zhang2020real,lu2023autonomous}. 
This topic has been studied for decades, with relevant areas including depth estimation~\cite{ranftl2021vision, yang2024depth}, multi-view stereo (MVS)~\cite{yao2018mvsnet, gu2020cascade}, novel view synthesis (NVS)~\cite{mildenhall2021nerf, guo2024uc, guo2024free}, Structure-from-Motion (SfM)~\cite{schonberger2016structure}, and Simultaneous localization and mapping (SLAM)~\cite{campos2021orb, teed2021droid}.

However, estimating scale-consistent 3D structures from dynamic monocular surgical video remains a challenging and ill-posed problem. This challenge arises from sparse features, the lack of multi-view constraints, and the complexity of surgical environment, which involves factors such as illumination variance, textureless surfaces, motion blur, and dynamic deformations from surgical interventions.
Traditional methods~\cite{recasens2021endo, schonberger2016structure}, which are developed under the assumption of rigid scenes, struggle to extract reliable features and match correspondences across frames in such dynamic environments.
Although recent monocular depth foundation models~\cite{ranftl2021vision, yang2024depth}
have made significant progress, they degrade when applied to surgical scenes and fail to predict accurate relative geometry. 
Some methods attempt to transfer general-domain models to surgical video, but they either require prior information (e.g., camera parameters or instrument models)~\cite{shao2022self, wei2024enhanced, wei2024scale}, or adopt an SfM-like multi-stage pipeline to learn both motion and geometry by estimating correspondences, camera poses and intrinsics for higher relative scale consistency~\cite{shao2022self, cui2024endodac}. 
Moreover, such SfM-like multi-stage pipeline will accumulate errors in every stage or require offline optimization, leading to sub-optimal accuracy and consistency. 

\begin{figure*}[t]
\centering
\includegraphics[width=1.0\linewidth]{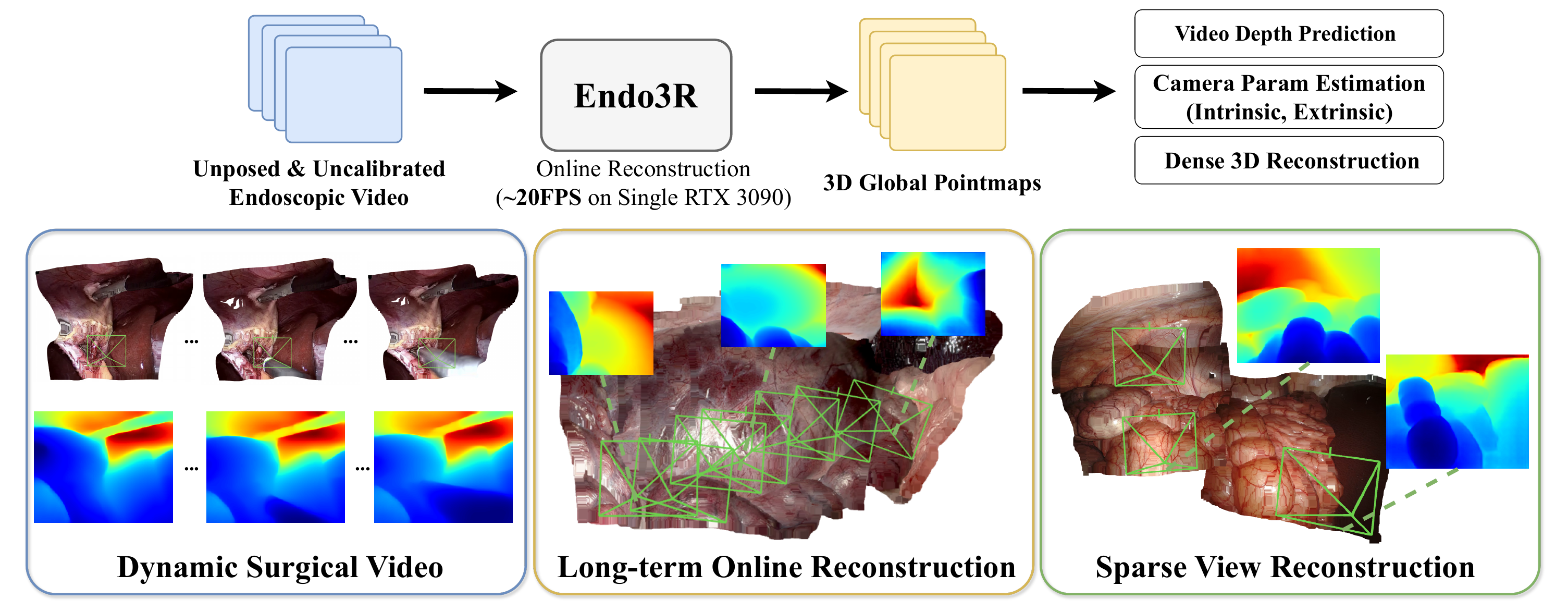}
\vspace{-2.5em}
\caption{Given monocular surgical video as input, our Endo3R allows feed-forward output of global pointmaps, scale-consistent depth, and camera parameters.
}
\vspace{-10pt}
\label{fig: teaser}
\end{figure*}

In this paper, we address these challenges and present \textbf{Endo3R}, a unified 3D surgical foundation model for online scale-consistent reconstruction from monocular endoscopic video \textit{without any prior information or extra optimization}, predicting globally aligned pointmaps, scale-consistent video depth, camera poses and intrinsics, as shown in Fig.~\ref{fig: teaser}. The key contribution of our method is devising an uncertainty-aware dual memory mechanism to expand the pairwise reconstruction ability from DUSt3R~\cite{wang2024dust3r} to long-term incremental dynamic reconstruction, by capturing both short-term dynamics and long-term spatial memory. We employ a memory encoder to save history tokens as memory keys and values, retrieving the relative information by cross attention. We measure the uncertainty of tokens by calculating the Sampson distance and filter out tokens with high uncertainty. Regarding the lack of training datasets, we introduce self-supervised training scheme for data without ground-truth poses and depths. Namely, a dynamics-aware flow loss is designed to enforce the cross-frame temporal consistency.

Our contribution is summarized as follows: $\mathbf{1)}$ We present Endo3R, a 3D surgical foundation model to enable real-time reconstruction from monocular video, unifying the prediction of globally aligned pointmaps, scale-consistent video depth, camera poses, and intrinsics. $\mathbf{2)}$ We present an uncertainty-aware dual memory mechanism to enable long-term online dynamic reconstruction. $\mathbf{3)}$ A self-supervised scheme is introduced to allow for scaling to more surgical datasets without ground truth. $\mathbf{4)}$ Experimental results demonstrate our superior performance in video depth estimation and pose estimation with online efficiency.

\section{Methodology}

In this paper, we aim to build a unified framework to solve online 3D reconstruction from endoscopic video, by adapting the static pairwise reconstruction from DUSt3R to long-term endoscopic videos. To enhance the robustness in long-term learning, our main insight is to enable incremental online reconstruction by an uncertainty-aware dual memory mechanism, predicting globally aligned pointmaps, temporally consistent video depth, camera poses, and intrinsics. Due to the scarcity of surgical datasets, we further employ a hybrid training mechanism and devise a flow-guided self-supervised learning to help scale up our network to more surgical datasets with different scenes. 

\begin{figure*}[t]
\centering
\includegraphics[width=1.0\linewidth]{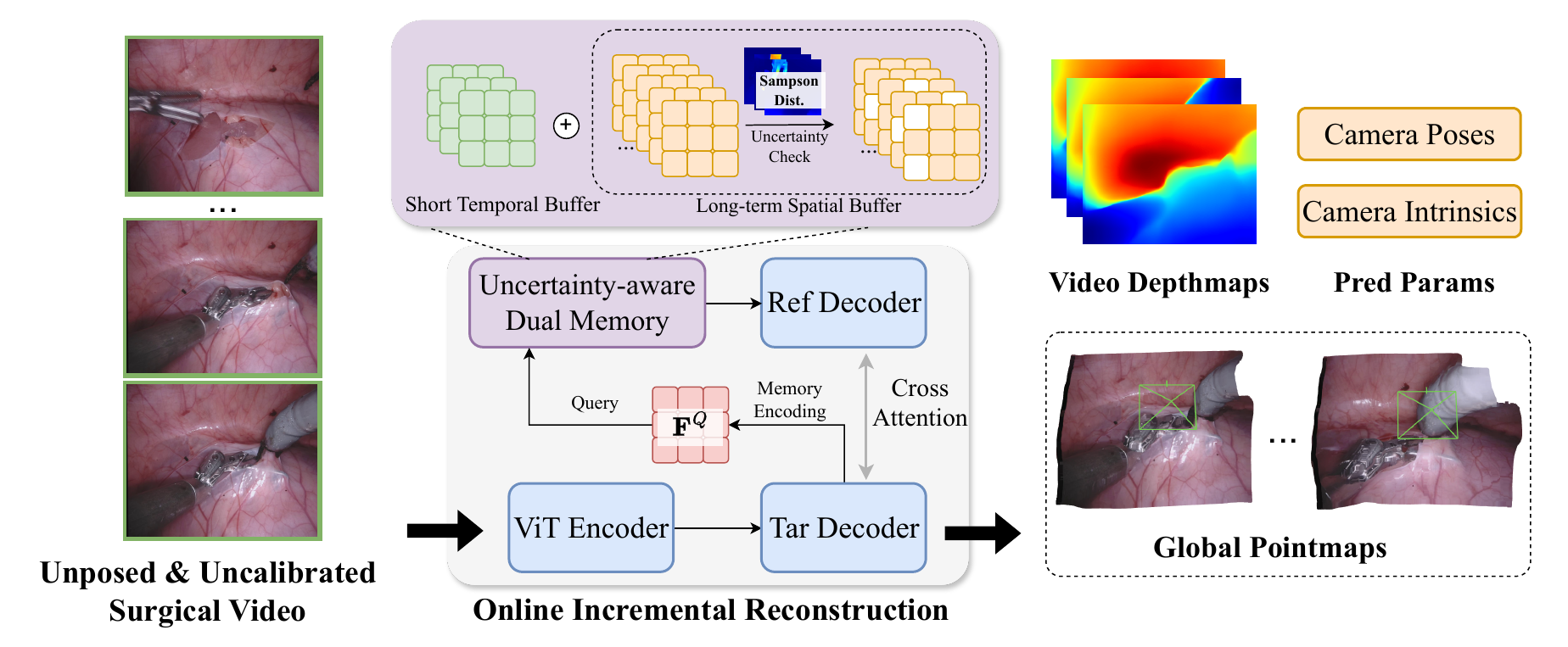}
\vspace{-2.5em}
\caption{\textbf{Overview of Endo3R}. Given monocular surgical video as input, we present a 3D surgical foundation model to enable online reconstruction from video. 
}
\label{fig: method}
\end{figure*}
As shown in Fig.~\ref{fig: method}, given a sequence of images $\{ \bI_i\}_{i=1}^N \in \nR^{W \times H \times 3}$ as input, our goal is to train a network $\mathcal{F}$ to output the corresponding pointmaps $\{\bX_{i, 1} \}_{i=1}^N \in \nR^{W \times H \times 3}$, confidence maps $\{\bC_{i, 1} \}_{i=1}^N \in \nR^{W \times H }$ in the coordinate frame of $t=1$ (Sec.~\ref{sec: network}). To tackle the highly dynamic long-term surgical video, we incorporate an uncertainty-aware spatial-temporal memory to preserve history tokens, capturing both short-term dynamics and long-term spatial consistency (Sec.~\ref{sec: memory}). Regarding the scarcity of surgical datasets with groundtruth, we design a self-supervised flow loss by decomposing the optical flow into scene flow and camera projection flow, enforcing the scale continuity and pose smoothness between consecutive frames (Sec.~\ref{sec: training}).
\subsection{Network architecture}\label{sec: network}
\noindent\textbf{Encoding. }Given a sequence of images as input, our network first encode every frame $\bI_i$ into tokens $\bF_i$ with a ViT encoder~\cite{dosovitskiy2020image}: $\bF_i = \text{Encoder}(\bI_i)$.

\noindent\textbf{Memory retrieval.}
Unlike~\cite{wang2024dust3r} conducting pairwise prediction, we enable the incremental reconstruction by incorporating an attention-based memory mechanism inspired by Spann3R~\cite{wang20243d}. The memory bank stores the historical key features and value features. 
Given every new frame, we leverage a previous query feature $\bF_{t-1}^Q$ to retrieve relative contexts from the memory bank to output the fused tokens $\bF_{t-1}^G$:
\begin{equation}
    \bF_{i-1}^G = \text{Softmax}(\frac{\bF_{i-1}^Q (\bF^K)^\text{T}}{\sqrt{C}}) \bF^V + \bF_{i-1}^Q,
\end{equation}
where $\bF^K$ and $\bF^V$ are key and value features saved in the memory bank.

\noindent\textbf{Decoding. }After encoding, two transformer decoders sequentially perform self-attention and cross-attention on both encoded feature $\bF_t$ and fused feature $\bF_{t-1}^G$ to predict the 3D geometry: ${\bF'}_i, {\bF'}_{i-1}^{G} = \text{Decoder}(\bF_{i}, \bF_{i-1}^G)$,
where ${\bF'}_i$ and ${\bF'}_{i-1}^{G}$ denote the features after the cross-view interaction. 

\noindent\textbf{Regression Head. }After decoding, the 3D representations are predicted from the decoded features. Following~\cite{wang2024dust3r}, we employ DPT~\cite{ranftl2021vision} head to predict the 3D pointmap and associated confidence map. We compute the camera pose $\hat{\bT}_{i, 1}$ based on PnP. Then the depth $\hat{\bD}_i$ could be estimated by transforming the global pointmap to the local coordinate with $\hat{\bT}_{i, 1}$:
\begin{align}
    \hat{\bX}_{i, 1}, \hat{\bC}_{i, 1} &= \text{Head}_{\text{output}}({\bF'}_i), \\
    \hat{\bD}_i &= (\hat{\bT}_{i, 1} \hat{\bX}_{i, 1})_z.
\end{align}

\subsection{Uncertainty-aware Dual Memory}\label{sec: memory}

\noindent\textbf{Dual Memory. }To extend~\cite{wang2024dust3r} to sequential reconstruction, we introduce an uncertainty-aware dual memory mechanism consisting of a long-term spatial buffer and short-term temporal buffer. Namely, global keyframe tokens and stable 3D information are stored in the long-term spatial buffer, maintaining spatial consistency over time. The short-term temporal buffer stores tokens from the recent frames, ensuring temporal consistency across consecutive frames.

\noindent\textbf{Memory Encoding. }
At the end of each step, the decoded feature ${\bF'}_i$ and encoded feature $\bF_i$ are used to generate the query feature for the next step. The information of the current frame is preserved in the short-term temporal buffer of the memory bank as key and value features. When more frames come in, the older memory keys and values will be moved to the long-term spatial buffer.

\noindent\textbf{Uncertainty Check. }
Unlike static reconstruction, dynamic surgical scenes present additional challenges, e.g., non-rigid tissue deformations, surgical instruments frequently appearing and disappearing, and occlusions due to sudden camera movements or interactions with anatomical structures. Therefore, we aim to filter the memory bank to eliminate the 3D information of transient objects and occlusions, to enhance the global 3D consistency and robustness for the new incoming frames. To filter out dynamic tokens and disturbances, we use the Sampson distance to assess the reliability of the tokens stored in the long-term spatial memory. We follow ~\cite{guo2024free} to leverage the optical flow $\bO_{i \rightarrow i+1}$ to assess the epipolar geometry with the estimated poses $\hat{\bT}_i$ and $\hat{\bT}_{i+1}$. Therefore, given every encoded memory as input, the tokens with high Sampson distance (i.e. larger than threshold $\beta$) indicate unreliable matches and will be eliminated from the memory bank $\bF^K$ and $\bF^Q$. For long sequence inference, we leverage confidence map $\bC$ to select top $K$ tokens in the memory bank and prune the others.

\subsection{Self-supervised Losses}\label{sec: training} 
Despite the success of DUSt3R-related methods, they require supervised training on large-scale datasets with both GT depth and poses.
However, in surgical scenes, there are limited datasets containing the GT depth and poses, which hinders the training for diverse scenes or surgeries with monocular videos only. 
To address this problem, we propose a self-supervised training scheme that enables training on datasets without full labels.

\noindent\textbf{Dynamics-aware Flow Loss. }
Previous monocular depth estimation methods~\cite{chen2025video, yang2024depth, budd2024transferring} enforce temporal consistency by minimizing the difference between the flow-warped depth $\hat{\bD}_i$ and $\hat{\bD}_{i+1}$, assuming that the depths of corresponding points remain stationary. However, this assumption does not hold in real-world surgical scenes, which feature dynamic instruments and deformable tissues.

\begin{figure*}[t]
\centering
\includegraphics[width=1.0\linewidth]{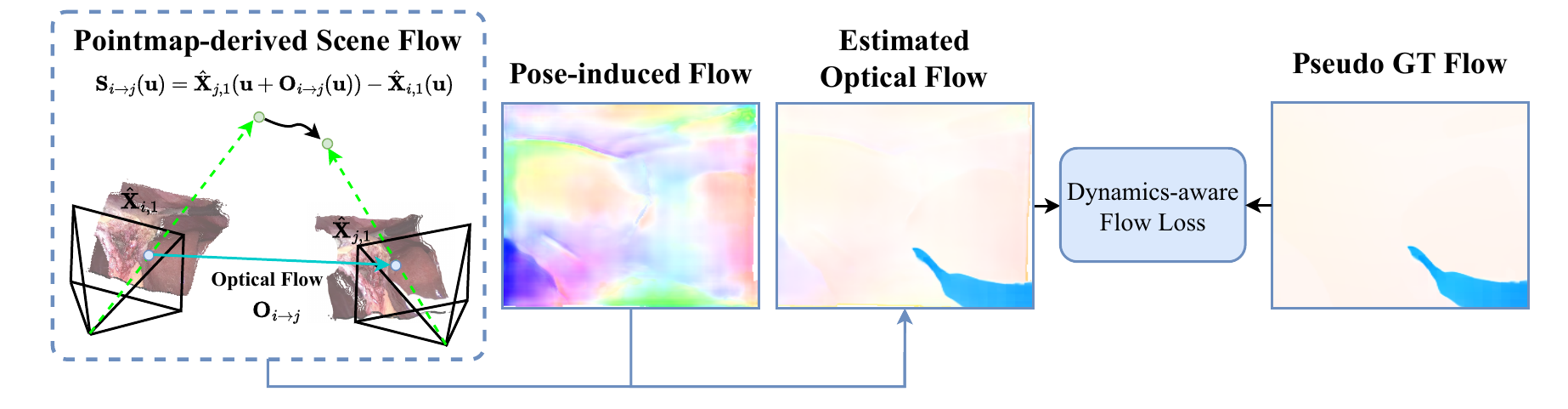}
\vspace{-2.5em}
\caption{\textbf{Illustration of dynamics-aware flow loss} for self-supervised training to achieve enhanced temporal consistency.}
\label{fig: method_2}
\end{figure*}

To address this limitation, as shown in Fig.~\ref{fig: method_2}, we propose a dynamics-aware flow loss that eliminates the stationary assumption by decoupling optical flow into pose-induced motion and pointmap-derived scene flow. 
Specifically, given the input image sequences, we first compute the forward optical flow as $\bO_{i \rightarrow j}$ with off-the-shelf model~\cite{teed2020raft}. Optical flow captures 2D motion of pixels between frames, encompassing both camera motion and scene flow. To calculate the scene flow between frame $i$ and $j$, we leverage optical flow to find correspondences between the pointmaps $\hat{\bX}_{i, 1}$ and $\hat{\bX}_{j, 1}$, the scene flow is calculated as:
\begin{equation}
    \hat{\bS}_{i \rightarrow j}(\bu) = \hat{\bX}_{j, 1}(\bu + \bO_{i \rightarrow j}(\bu)) - \hat{\bX}_{i, 1}(\bu),
\end{equation}
where $\bu$ is the homogeneous 2D coordinate. We restrict the computation to the valid region and define $\bu' = \{\bu | 0< \bu+ \bO_{i \rightarrow j}(\bu)<(H, W)\}$. Then the estimated optical flow $ \hat{\bold{f}}_{i \rightarrow j}$ could be calculated by combining the scene flow with pose-induced flow as:
\begin{equation}
    \hat{\bold{f}}_{i \rightarrow j}(\bu') = \hat{\bK} \hat{\bT}_{j, 1} (\hat{\bX}_{i, 1}(\bu’) + \hat{\bS}_{i \rightarrow j}(\bu’)) - \bu’,
\end{equation}
where $\hat\bK$ denotes the estimated intrinsic by solving a simple optimization following~\cite{wang2024dust3r}. The dynamics-aware flow loss can be written as:
\begin{equation}
    \mathcal{L}_{\text{Dflow}}^{i\rightarrow j} = \parallel \hat{\bold{f}}_{i \rightarrow j}(\bu') -\bO_{i \rightarrow j} (\bu')\parallel_1.
\end{equation}
Based on $\mathcal{L}_{\text{Dflow}}^{i\rightarrow j}$, we avoid the need of camera pose and depth for training.

\noindent\textbf{Monocular Depth Loss.}
For datasets without either GT depth or pose, we use off-the-shelf video depth model~\cite{chen2025video} to obtain the monocular depth and adopt a scale-invariant depth loss in Midas~\cite{ranftl2020towards} to supervise the predicted depth $\hat\bD$. We first calculate the shift and scale by least square to align $\bD$ to $\hat{\bD}$ and obtain $\Tilde{\bD}$, then minimize the $\mathcal{L}_2$ loss and gradient loss as follows:
\begin{equation}
    \mathcal{L}_{\text{dep}} = \mathcal{L}_{2} + \mathcal{L}_{\text{smooth}} = \frac{1}{M} \parallel \Tilde{\bD} - \bD \parallel^2_2 + \frac{1}{M} \sum_{k=1}^{K}\sum_{i=1}^M ( \mid \nabla_x \bR_i^k + \nabla_y \bR_i^k \mid),
\end{equation}
where $\bR_i$ denotes the difference between $\Tilde{\bD}$ and $\hat{\bD}$ with scale level $K=4$, $M$ denotes the total pixels of image. 
\begin{table}[tbp]
\scriptsize
\centering
\caption{\textbf{Quantitative comparison with depth estimation methods}.
}
\vspace{-1em}
\label{tab: results}
\resizebox{0.95\linewidth}{!}{%
\begin{tabular}{@{}l|l|ccccc|c@{}}
\toprule
&\textbf{Methods}&\textbf{Abs Rel}$\downarrow$ & \textbf{Sq Rel}$\downarrow$ & \textbf{RMSE}$\downarrow$ & \textbf{RMSE log}$\downarrow$&\textbf{$\boldsymbol{\delta}<\text{1.25}\uparrow$}  & \textbf{FPS}$\uparrow$
\\ 
\midrule
 \multirow{10}{*}{\rotatebox{90}{SCARED}} 
 & Monodepth2~\cite{godard2019digging} & 0.432& 3.548&4.704&0.431&0.425&22.05 \\
 & Endo-SfM~\cite{ozyoruk2021endoslam} & 0.241& 0.865&2.286&0.267&0.585&7.33 \\
 & AF-SfM~\cite{shao2022self} & 0.257& 0.960&2.162&0.291&0.573&3.17\\ 
& EndoDAC~\cite{cui2024endodac} & 0.242& 0.934&2.014&0.275&0.584&31.79\\
& Transfer~\cite{budd2024transferring} &0.297& 1.207&2.561&0.319&0.561&9.37\\
& DA-V2~\cite{yang2025depth} & 0.313& 1.425&2.839&0.453&0.508&4.18\\
 &VDA~\cite{chen2025video} & 0.291& 1.186&2.447&0.296&0.647&6.86\\
& Endo DM~\cite{recasens2021endo}& 0.203& 0.651&2.063&0.245&0.612&14.58 \\
& Monst3R~\cite{zhang2024monst3r}& 0.198& 0.539&1.965&0.234&0.626&18.68 \\
&\textbf{Endo3R(Ours)} & \textbf{0.124} & \textbf{0.227} & \textbf{1.209} & \textbf{0.135} &  \textbf{0.839} &19.17\\
\midrule
\multirow{10}{*}{\rotatebox{90}{Hamlyn}}
& Monodepth2~\cite{godard2019digging} & 0.379& 9.318&20.472&0.403&0.439&22.05 \\
& Endo-SfM~\cite{ozyoruk2021endoslam} & 0.252& 4.335&14.430&0.268&0.628&7.33 \\
&AF-SfM~\cite{shao2022self} &0.286& 5.715&15.895&0.301&0.508&3.17 \\ 
&EndoDAC~\cite{cui2024endodac} &0.275& 5.557&15.669&0.288&0.519&31.79 \\ 
& Transfer~\cite{budd2024transferring} & 0.281& 5.790&15.936&0.312&0.504&9.37\\
& DA-V2~\cite{yang2025depth} & 0.334& 7.713&19.548&0.362&0.461&4.18\\
 &VDA~\cite{chen2025video} & 0.315& 7.492&19.231&0.347&0.476&6.86\\
& Endo DM~\cite{recasens2021endo} & 0.216& 4.639&14.799&0.273&0.619&14.58\\
& Monst3R~\cite{zhang2024monst3r}& 0.198& 4.193&15.221&0.241&0.645&18.68 \\
&\textbf{Endo3R(Ours)} &\textbf{0.170} &\textbf{3.139} &\textbf{11.569}&\textbf{0.196}&\textbf{0.707}  &19.17\\
 \bottomrule 
\end{tabular}
}
\end{table}

\subsection{Training and Inference}
\noindent\textbf{Total Loss.}
Our total loss for training Endo3R is as follows:
\begin{equation}
    \mathcal{L}_{all} = \lambda_1 \mathcal{L}_{\text{Dflow}} +  \lambda_2 \mathcal{L}_{\text{dep}} + \lambda_3 \mathcal{L}_{\text{conf}},
\end{equation}
 where $\mathcal{L}_{\text{conf}}$ denotes the confidence-aware regression loss to supervise the pointmaps following~\cite{wang2024dust3r}, $\lambda_1, \lambda_2, \lambda_3$ denote the weights for losses.
 \begin{figure*}[t]
\centering
\includegraphics[width=1.0\linewidth]{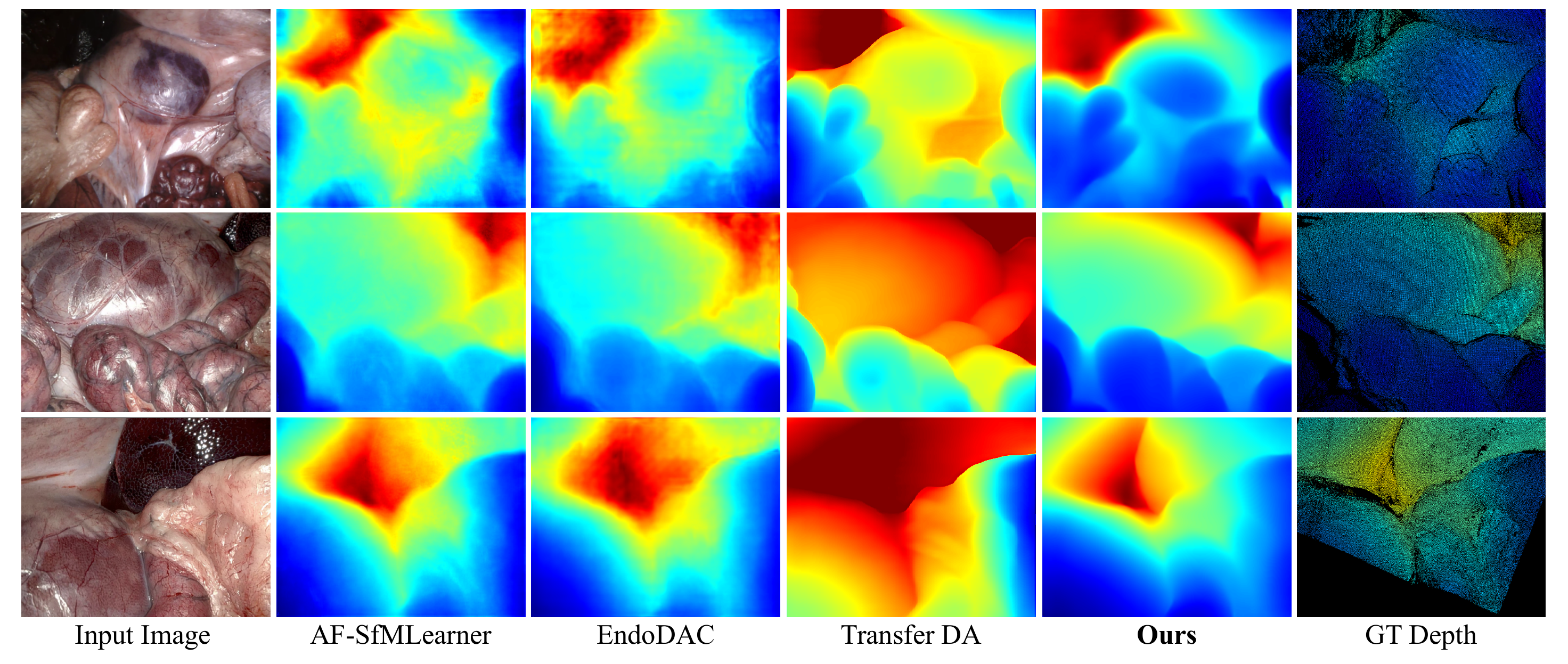}
\vspace{-2.5em}
\caption{\textbf{Qualitative results} of monocular depth estimation.}
\label{fig: qual}
\end{figure*}

\begin{table}[t!]
\def\arraystretch{0.9}
\begin{minipage}{0.48\textwidth}
    \centering
    \small
\caption{\textbf{Comparison of Pose Estimation} on the SCARED Dataset.}
\vspace{-0.6em}
\label{tab:comp_pose}
\resizebox{0.95\linewidth}{!}{%
    \begin{tabular}{lccc}
        \toprule
         \multirow{1}{*}{\textbf{Method}} &
         \textbf{ATE} $\downarrow$ & \textbf{RPE$_r$} $ \downarrow$& \textbf{RPE$_t$} $\downarrow$  \\
         \midrule
         Endo-SfM~\cite{ozyoruk2021endoslam} &0.157&0.252&0.259\\
         AF-SfM~\cite{shao2022self}&0.125&0.235&0.241 \\
         EndoDAC~\cite{cui2024endodac} &0.124&0.223&0.233 \\
         Robust~\cite{hayoz2023learning} &0.131&0.241&0.245\\
         \textbf{Endo3R(Ours)} &\textbf{0.112} & \textbf{0.201} & \textbf{0.228}\\
        \bottomrule
    \end{tabular}
}
\end{minipage}%
\hfill
\begin{minipage}{0.48\textwidth}
     \centering
    \small
\caption{\textbf{Ablation study} of Endo3R for different components.
}
\vspace{-0.6em}
\label{tab: ablation}
\resizebox{1.0\linewidth}{!}{%
\begin{tabular}{cccc}
\toprule
\textbf{Setting} &\textbf{Abs Rel}$\downarrow$ & \textbf{RMSE}$\downarrow$&\textbf{$\boldsymbol{\delta}<\text{1.25}\uparrow$} 
\\
\midrule
Baseline  & 0.198 & 1.965 & 0.626\\
 w/ Uncertain. & 0.165 & 1.654 & 0.720   \\
w/  $\mathcal{L}_{\text{dep}}$ & 0.153 & 1.486 & 0.772 \\
w/ $\mathcal{L}_{\text{Dflow}}$ &\textbf{0.124} & \textbf{1.209} & \textbf{0.839}  \\
\bottomrule
\end{tabular}
}
\end{minipage}
\end{table}

\begin{figure*}[t]
\centering
\includegraphics[width=0.85\linewidth]{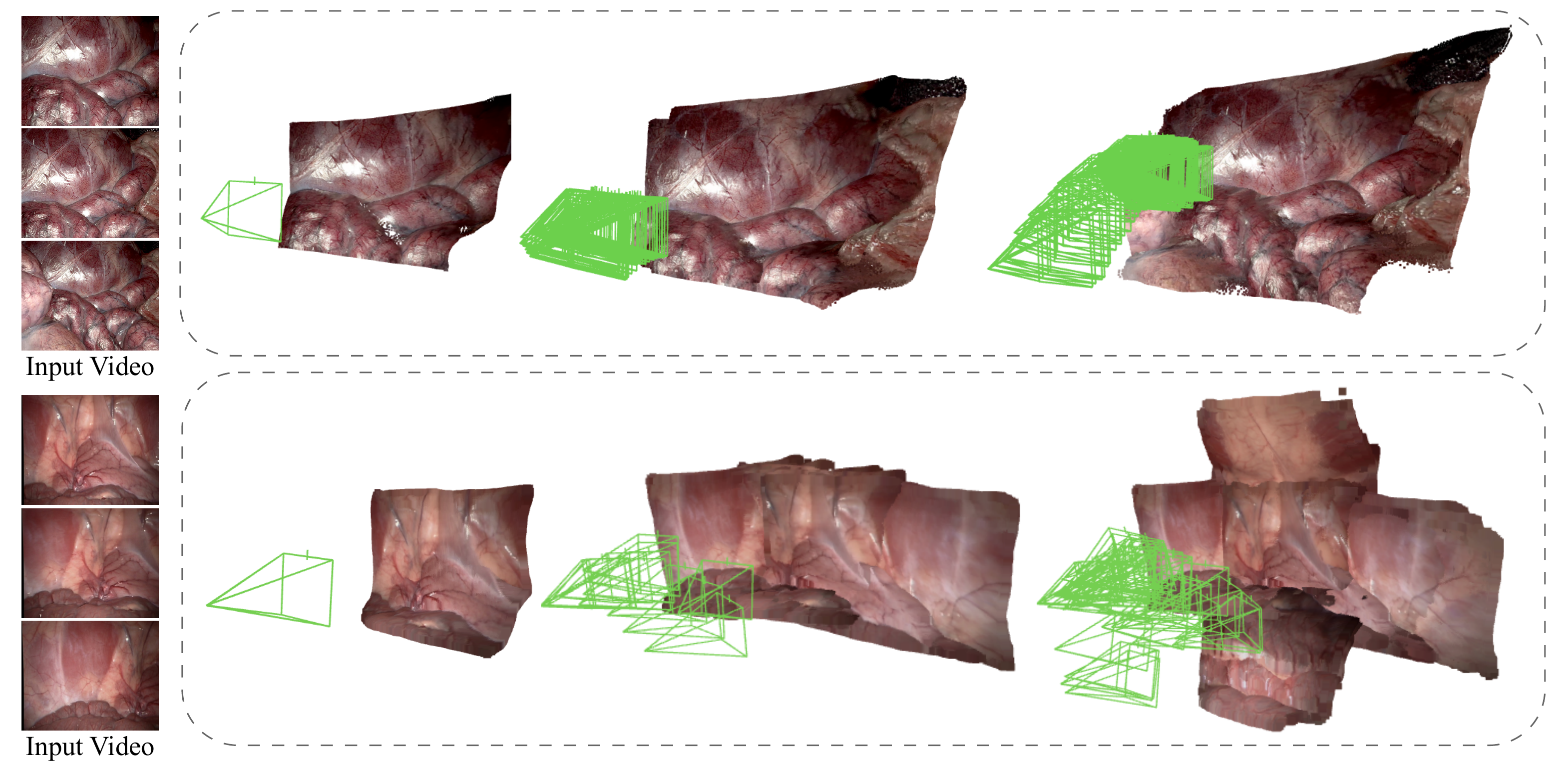}
\vspace{-1em}
\caption{\textbf{Qualitative results} of Online 3D Reconstruction.}
\label{fig: qual2}
\end{figure*}
\section{Experiments}
\subsection{Implementation Details}

\noindent \textbf{Training Datasets. }We train our Endo3R with a mixture of datasets, with four datasets containing GT/Stereo depth and pose (SCARED~\cite{allan2021stereo}, StereoMIS~\cite{hayoz2023learning}, C3VD~\cite{bobrow2023}, and Endomapper~\cite{azagra2023endomapper}), four datasets without GT data (AutoLaparo~\cite{wang2022autolaparo}, Cholec80~\cite{twinanda2016endonet}, EndoVis17~\cite{allan20192017}, and EndoVis18~\cite{allan20202018}). Specifically, we conduct stereo rectify for SCARED~\cite{allan2021stereo} and StereoMIS~\cite{hayoz2023learning}, using StereoAnything~\cite{guo2024stereo} to calculate the stereo depth of left view for training. To evaluate the depth estimation, we evaluate our method on $320 \times 256$ resolution and follow the train and test split in SCARED~\cite{allan2021stereo}. To evaluate the generalization ability, we test on all 22 videos of unseen Hamlyn Dataset for cross-dataset zero-shot validation.

\noindent \textbf{Evaluation Metrics. }We compare Endo3R with state-of-the-art depth estimation methods. We follow~\cite{cui2024endodac} to use five metrics commonly used in monocular depth estimation: Abs Rel, Sq Rel, RMSE, RMSE log, $\delta < 1.25$. We also compare the inference FPS to compare the efficiency. To evaluate the pose accuracy, we perform a 5-frame pose evaluation and adopt Absolute Trajectory Error (ATE) and Relative Pose Error (RPE), including rotation RPE$_r$ and translation RPE$_t$. Note that the unit for RPE$_t$ and ATE is mm, and the unit for RPE$_r$ is degree.

\subsection{Experimental Results}
\noindent\textbf{Quantitative Comparison.} 
We evaluate our method and SOTA depth estimation methods on SCARED and Hamlyn datasets.
The results in Tab.~\ref{tab: results} reveal that our approach achieves a substantial improvement in depth estimation accuracy compared to existing methods, even without training on the Hamlyn Dataset. Notably, while delivering superior accuracy, our method maintains a competitive FPS rate to support online applications. We also report the pose estimation results on SCARED in Tab.~\ref{tab:comp_pose}. The results demonstrate that our method achieves the highest pose estimation accuracy. 

\noindent\textbf{Qualitative Comparison.}
The qualitative evaluation of our depth estimation is illustrated in Fig.~\ref{fig: qual}, demonstrating that Endo3R produces more precise depth maps with improved relative scale. Furthermore, Fig.~\ref{fig: qual2} presents our online 3D reconstruction results with pose estimation. The high-quality 3D reconstructions can be attributed to the superior depth and pose estimation accuracy. Please find more visualization results in the supplementary video.

\noindent\textbf{Ablation Study. }We set Monst3R~\cite{zhang2024monst3r} as baseline and conduct ablation studies on the different components of Endo3R. As reported in Tab.~\ref{tab: ablation}, it shows the effectiveness of each component with increasing performance.

\section{Conclusion}
We have presented Endo3R, a unified framework for online 3D reconstruction from uncalibrated surgical videos. By jointly learning depth, pose, and scene geometry in a single stage, our method eliminates the need for multi-stage pipelines or offline optimization. The proposed uncertainty-aware memory mechanism and self-supervised learning paradigm effectively address the challenges of dynamic surgical scenes and limited annotated data. Experimental results demonstrate the framework's robustness and efficiency, showcasing its potential for practical surgical applications. This work provides a foundation for future research in real-time surgical scene understanding and computer-assisted intervention.

\begin{credits}
\subsubsection{\ackname} 
This work is supported in part by the Shenzhen Portion of Shenzhen-Hong Kong Science and Technology Innovation Cooperation Zone under HZQB-KCZYB-20200089, in part of the HK RGC under AoE/E-407/24-N, Grant 14218322, and Grant 14207320, in part by the Hong Kong Centre for Logistics Robotics, in part by the Multi-Scale Medical Robotics Centre, InnoHK, and in part by the VC Fund 4930745 of the CUHK T Stone Robotics Institute.
\end{credits}

\bibliographystyle{splncs04}
\bibliography{egbib}

\end{document}